\documentclass[a4paper,fleqn]{cas-dc}

\usepackage[authoryear,longnamesfirst]{natbib}

\def\tsc#1{\csdef{#1}{\textsc{\lowercase{#1}}\xspace}}
\tsc{WGM}
\tsc{QE}
\tsc{EP}
\tsc{PMS}
\tsc{BEC}
\tsc{DE}

\begin{document}
\let\WriteBookmarks\relax
\def\floatpagepagefraction{1}
\def\textpagefraction{.001}

\shorttitle{E-ConvNeXt}

\shortauthors{F Wang et~al.}

\title [mode = title]{E-ConvNeXt: A Lightweight and Efficient ConvNeXt Variant with Cross-Stage Partial Connections}                      

\author[1]{Fang Wang}
[orcid=0000-0001-6170-0463]
\ead{fangwang@bipt.edu.cn}
\fnmark[1]

\affiliation[1]{organization={College of Information Engineering},
    addressline={Beijing Institute of Petrochemical Technology}, 
    city={Beijing},
    postcode={102617}, 
    country={People’s Republic of China}}

\author[1]{Huitao Li}[
orcid=0009-0005-4750-6273
]
\ead{2024520241@bipt.edu.cn}

\author[2]{Wenhan Chao}[
orcid=0000-0001-8028-5285
]
\ead{chaowenhan@buaa.edu.cn}
\affiliation[2]{organization={School of Computer Science and Engineering},
    addressline={Beihang University}, 
    city={Beijing},
    postcode={100191}, 
    country={People’s Republic of China}}

\author[1]{Zheng Zhuo}[
orcid=0000-0001-8028-5285
]
\ead{0020230022@bipt.edu.cn}
\cormark[1]

\author[1]{Yiran Ji}

\author[1]{Chang Peng}

\author[1]{Yupeng Sun}

\author[1]{jiyu zhang}

\cortext[cor1]{Corresponding author}

\begin{abstract}
Many high-performance networks were not designed with lightweight application scenarios in mind from the outset, which has greatly restricted their scope of application. This paper takes ConvNeXt as the research object and significantly reduces the parameter scale and network complexity of ConvNeXt by integrating the Cross Stage Partial Connections mechanism and a series of optimized designs. The new network is named E-ConvNeXt, which can maintain high accuracy performance under different complexity configurations. The three core innovations of E-ConvNeXt are : (1) integrating the Cross Stage Partial Network (CSPNet) with ConvNeXt and adjusting the network structure, which reduces the model's network complexity by up to 80\%; (2) Optimizing the Stem and Block structures to enhance the model's feature expression capability and operational efficiency; (3) Replacing Layer Scale with channel attention. Experimental validation on ImageNet classification demonstrates E-ConvNeXt's superior accuracy-efficiency balance: E-ConvNeXt-mini reaches 78.3\% Top-1 accuracy at  0.9GFLOPs.  E-ConvNeXt-small reaches 81.9\% Top-1 accuracy at 3.1GFLOPs.  Transfer learning tests on object detection tasks further confirm its generalization capability.
\end{abstract}


\maketitle

\section{Introduction}

Neural networks\cite{lecun1989backpropagation} have developed rapidly in the field of computer vision. The 2010s can be called the era of convolutional neural networks: AlexNet \cite{krizhevsky2012imagenet} pioneered the history of CNN-dominated computer vision; ResNet\cite{he2016deep}, with its wide adaptability to various downstream tasks, opened the chapter of image classification networks as backbone networks; and EfficientNet\cite{tan2019efficientnet} promoted neural networks toward a multi-variant development trend through compound scaling strategies. Entering the 2020s, the introduction of Transformer\cite{vaswani2017attention} has made the field of computer vision more open and diverse. Vision Transformer (ViT)\cite{dosovitskiy2020image} and Convolutional Neural Networks (ConvNets) learn from each other, giving birth to a series of new networks, whose development trends focus on higher accuracy, faster speed, and lower computational complexity.

ConvNeXt\cite{liu2022convnet} emerges as a representative of such hybrid networks that bridge the gap between traditional ConvNet and Transformers. As a key innovation in the 2020s, it inherits the strengths of ResNet's convolutional foundation while integrating core design ideas from Vision Transformer, aiming to combine the efficiency of pure convolutional architectures with the expressive power of Transformer-based models. ConvNeXt, based on ResNet, draws on the design concepts of Vision Transformer. It enlarges the convolution kernel to increase the effective receptive field, introduces depthwise convolution\cite{howard2017mobilenets} to control network parameters and network complexity, and simplifies the network architecture. It retains the efficient feature extraction capability of pure convolutional networks while absorbing the architectural advantages of Transformer. ConvNeXt outperforms Swin-Transformer\cite{liu2021swin} in image classification task\cite{russakovsky2015imagenet} and also demonstrates strong feature extraction capability in downstream tasks\cite{zhou2019semantic}\cite{lin2014microsoft}.

However, despite the excellent performance of ConvNeXt, it has shortcomings in lightweight adaptation\cite{howard2017mobilenets}\cite{sandler2018mobilenetv2}\cite{howard2019searching}.  Unlike EfficientNet-Lite\cite{tan2019efficientnet} and FasterNet-tiny\cite{chen2023run}, which are specifically designed for lightweight requirements, ConvNeXt lacks small-parameter variants. As a result, it is challenging to find suitable ConvNeXt variant that meet low-computation requirements in resource-constrained scenarios, such as mobile and embedded devices. 

ConvNeXt-tiny is the smallest variant of ConvNeXt, but the parameters of ConvNeXt-tiny are 28M, the network complexity of ConvNeXt-tiny is 4.5GFLOP. Compared with other lightweight networks, ConvNeXt still has a huge gap in terms of parameter count and model complexity, which greatly limits its application scenarios.

To address this critical gap, E-ConvNeXt is developed as a lightweight derivative of ConvNeXt. Explicitly engineered to mitigate ConvNeXt's inadequacies in lightweight support, E-ConvNeXt achieves significant reductions in network complexity and parameter through structural refinements and parameter recalibration, while preserving competitive performance metrics. This design enables seamless adaptation to resource-constrained scenarios, thereby extending the applicability of the ConvNeXt framework across a broader spectrum of practical use cases. The main contributions of E-ConvNeXt are as follows:

\begin{enumerate}

    \item This paper introduce CSPNet\cite{wang2020cspnet} into ConvNeXt and incorporate a transition hyperparameter into the CSP Stage, thereby significantly reducing the network complexity of ConvNeXt.

    \item We improve the model's operational efficiency and accuracy through a series of optimizations, specifically including: adjusting the Stem structure; replacing Layer Normalization with Batch Normalization; and using the channel attention mechanism to replace Layer Scale.
    
    \item Based on the aforementioned improvements, this paper proposes a new network named E-ConvNeXt. We conducted extensive experiments on ImageNet-1K to verify the effectiveness of E-ConvNeXt. Additionally, we transferred E-ConvNeXt to object detection tasks to validate its generalization capability.
\end{enumerate}

\section{Related Work}

\subsection{Lightweight Neural Network}

With the surging demand for mobile and edge computing scenarios, lightweight networks\cite{howard2017mobilenets}\cite{tan2019efficientnet} have gradually become an important research direction in the field of computer vision. Traditional ConvNets fail to meet the lightweight requirements in terms of parameters and network complexity.For example, the network complexity of ResNet-18 is 1.8GFLOPs, it‘s parameters are 11M. MobileNet\cite{howard2017mobilenets} replaces traditional convolution with Depthwise Separable Convolution, which significantly reduces parameters and network complexity of ConvNets. MobileNetV1\_x0\_25 network complexity is 0.07GFLOPs , it parameters are 0.460M. Early lightweight neural networks\cite{sandler2018mobilenetv2}\cite{howard2019searching}\cite{zhang2017shufflenetextremelyefficientconvolutional}\cite{zhang2018shufflenet} only focus on lightweight application scenarios and achieve good results in such scenarios, but they can't meet the requirements in high-performance scenarios.

EfficientNet\cite{tan2019efficientnet} can cater to multiple application scenarios by controlling the network's depth, width, and input image size.
EfficientNet has multiple variants. The network complexity of EfficientNet-B0 is 0.7GFLOPs, that of EfficientNetB7 is 72.3GFLOPs. EfficientNet has suitable variants to cope with lightweight application scenarios and high-performance application scenarios. Many ConvNets\cite{tan2021efficientnetv2}\cite{chen2023run} draw inspiration from EfficientNet and set up multiple variants to cope with different application scenarios.

However, many network do not consider lightweight application scenarios, and how to apply them to lightweight scenarios remains a relatively under-explored research direction.This paper takes ConvNeXt as the research object and combines CSPNet with ConvNeXt, significantly reducing the number of model parameters and network complexity while ensuring a certain level of accuracy.

\subsection{Cross Stage Partial Network (CSPNet)}

Cross Stage Partial Networks (CSPNet) respects the variability of
the gradients by integrating feature maps from the beginning and the end of a network stage.  
CSPNet is combined with many networks\cite{huang2017densely}\cite{xie2017aggregated}\cite{he2016deep}, it can reduce computations by 20\% with equivalent or even superior accuracy on the ImageNet dataset.

When ResNeXt-50 is transformed into CSPResNeXt-50, its network complexity decreases from 5.05 GFLOPs to 3.97 GFLOPs. When DenseNet-201\cite{huang2017densely} is transformed into CSPDenseNet-201, its network complexity drops from 4.35 GFLOPs to 3.05 GFLOPs. However, even after a 20\% reduction in network complexity, these networks still have a huge gap compared with lightweight networks such as MobileNet and EfficientNet, and thus cannot meet the lightweight requirements.This paper significantly reduces the network complexity and parameters of model by adjusting the structure of CSPNet. Meanwhile, it can  maintain good performance.

\subsection{Channel Attention}

Channel attention mechanisms aim to adaptively recalibrate feature channel weights by emphasizing discriminative channels and suppressing irrelevant ones, thereby enhancing model representational capacity. Pioneered by SENet\cite{hu2018squeeze}, this paradigm introduced the squeeze-and-excitation (SE) block, which aggregates global spatial information via global average pooling (GAP) in the squeeze module and models cross-channel dependencies using multi-layer perceptrons (MLPs) in the excitation module. 

Subsequent works have refined this framework: ECANet \cite{wang2020eca} replaced MLPs with 1D convolutions to capture local cross-channel interactions without dimensionality reduction, reducing computational complexity while preserving performance. GSoP-Net \cite{gao2019global} improved the squeeze module by incorporating global second-order pooling to model high-order statistical relationships between channels, enhancing global information capture. SRM\cite{lee2019srm} combined style pooling (mean and standard deviation) with lightweight channel-wise fully-connected layers, balancing efficiency and expressiveness. FcaNet\cite{qin2021fcanet} reinterpreted GAP through the lens of discrete cosine transforms (DCT), leveraging multi-spectral information to enrich global feature representation. 

ConvNeXt uses Layer Scale to help stabilize network training. This paper will explore how to use attention mechanisms to replace Layer Scale to further enhance the model's feature extraction capability.

\section{The proposed E-ConvNeXt}

In this section, we provide a trajectory going from the original ConvNeXt to our E-ConvNeXt. Figure \ref{fig:jiagoutu} shows the ideas and effects of each improvement step. To ensure fairness, we use ConvNeXt-tiny as the baseline and maintain the training strategy for all steps. All steps are implemented based on this trajectory and baseline setting, with specific improvements as follows:

\begin{figure} 
	\centering
		\includegraphics[width=1\linewidth]{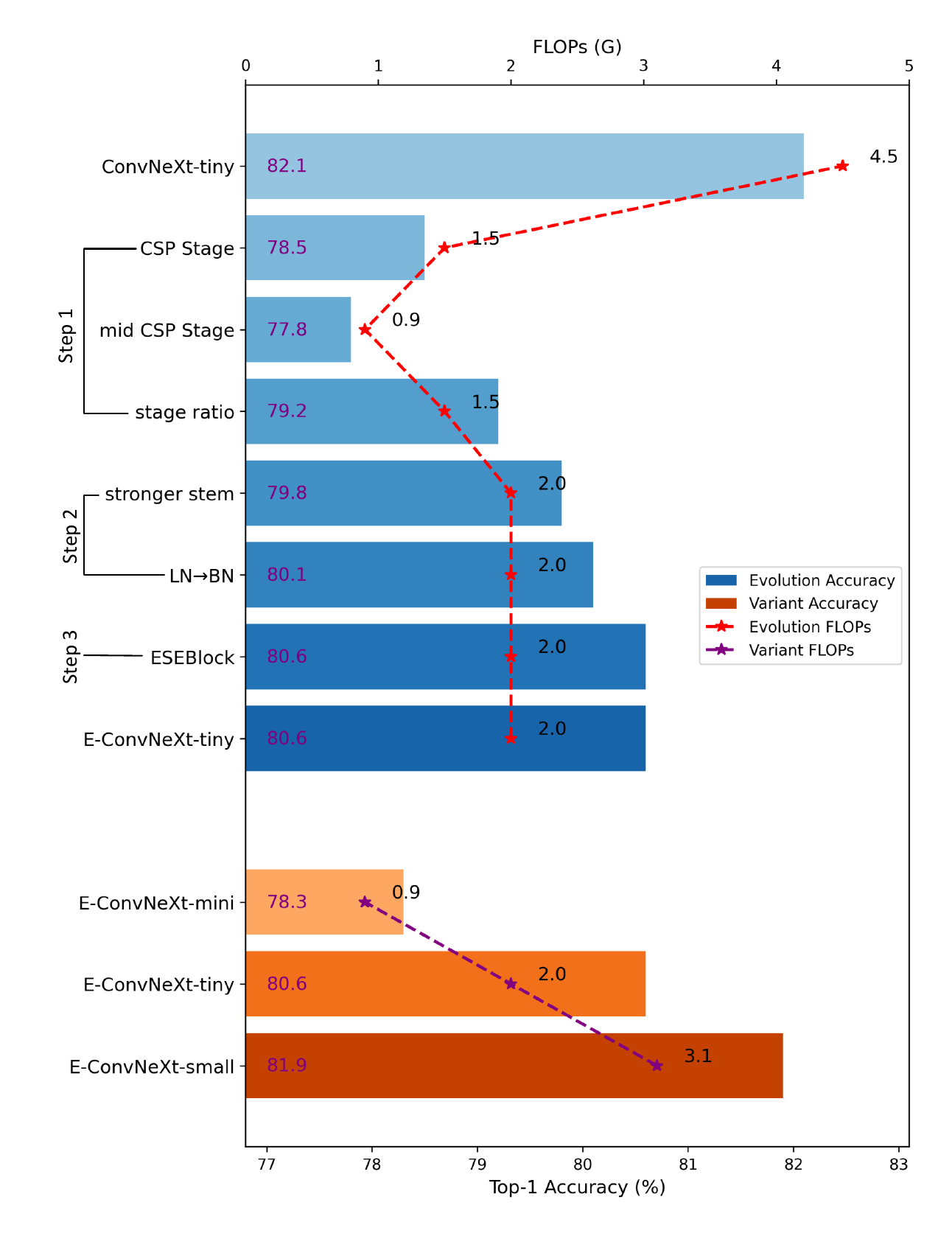}
	\caption{ \textbf{E-ConvNeXt Evolution Figure} shows the steps from ConvNeXt to E-ConvNeXt. The bar chart represents the Top-1 accuracy on ImageNet-1K, and the line chart shows the changes in model FLOPs.}
	\label{fig:jiagoutu}
\end{figure}

\begin{enumerate}
\itemsep=1pt
\item Combine CSPNet with ConvNeXt: we combine CSPNet with ConvNeXt and improve the CSP Module through the following three steps: 1) CSP Stage; 2) Introducing transition hyperparameter; 3) Stage ratio. As shown in Figure \ref{fig:jiagoutu}, this step significantly reduces network complexity. The specific details will be described in section \ref{chap:CSPNet_with_ConvNeXt};

\item  Optimization of the ConvNeXt structure: 1) A stepped stem is proposed to alleviate the problem of information loss caused by continuous downsampling. 2) replacing Layer Normalization\cite{bao2021beit} (LN) with Batch Normalization\cite{ioffe2017batch} (BN): BN is used to replace LN tor further improve efficiency.

\item Introducing channel attention: We solved the problem that adding channel attention to ConvNeXt would cause training collapse, and added the Effective Squeeze-and-Excitation Block\citep{lee2020centermask} (ESE Block) to the network.

\end{enumerate}

As shown in Figure \ref{fig:jiagoutu}, step 1 can significantly reduce  the network complexity, and step 2 and 3 enhance the model's efficiency and accuracy. Finally, we get E-ConvNeXt-tiny from ConvNeXt-tiny. Building on E-ConvNeXt-tiny, we further extend it into two variants: E-ConvNeXt-mini and E-ConvNeXt-small. Among them, E-ConvNeXt-mini network complexity is 0.9 GFLOPs and Top-1 accuracy is 78.3\%; E-ConvNeXt-tiny network complexity is 2.0 GFLOPs and Top-1 accuracy is 80.6\%; E-ConvNeXt-small network complexity is 3.1 FLOPs and  Top-1 accuracy is 81.9\%.

\subsection{Combine CSPNet with ConvNeXt}\label{chap:CSPNet_with_ConvNeXt}

\subsubsection {CSP Stage}

\begin{figure} 
	\centering
		\includegraphics[width=1\linewidth]{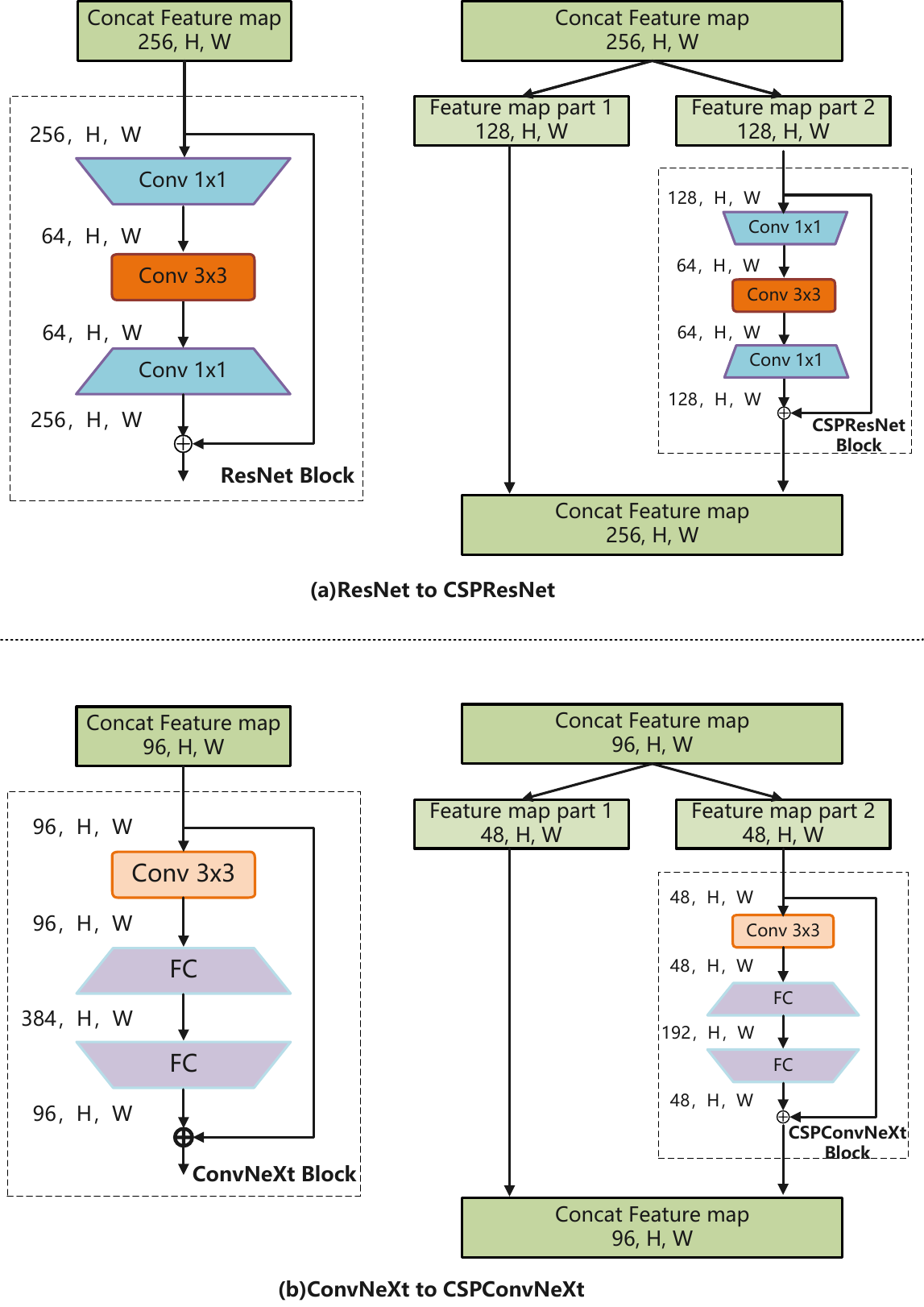}
	\caption{\textbf{CSPConvNets demonstration figure} shows how CSPNet transforms ResNet into CSPResNet, and how we transform ConvNeXt into CSPConvNeXt.}
	\label{fig:CSPduibi}
\end{figure}

CSPNet reduces parameters and FLOPs while maintaining accuracy by adjusting the channel numbers in some convolutions.
As shown in Figure \ref{fig:CSPduibi}, CSPResNet splits the feature map into two feature maps. The channel numbers of each new map are half of the original ones. One feature map passes through the blocks and the other does not. Then the two parts are merging together. Compared to ResNet, CSPResNet only changes the input channel and output channel of the 1x1 convolution from 256 to 128.The formula for calculating the convolution FLOPs is
\begin{equation}
FLOPs=C_{in} \times H_{out} \times W_{out} \times K^2 \times C_{out},
  \label{eq:FLOPs_Conv}
\end{equation}
The FLOPs of ResNet Block are
\begin{align*}
    ((256 \times 56 \times 56 \times 1^2 \times 64)\approx 51,4M)+ \\
    ((64 \times 56 \times 56 \times 3^2 \times 64) \approx 116M)+ \\
     ((64 \times 56 \times 56 \times 1^2 \times 256)\approx51.4M )\approx 218.8M.
  \label{eq:FLOPs_ResConv}
\end{align*}
The FLOPs of CSPResNet Block are
\begin{align*}
    ((128 \times 56 \times 56 \times 1^2 \times 64)\approx25.7M)+ \\
    ((64 \times 56 \times 56 \times 3^2 \times 64)\approx 116M)+ \\
     ((64 \times 56 \times 56 \times 1^2 \times 128)\approx 25.7M) \approx 167.4M.
\end{align*}
Compared with ResNet, CSPResNet reduced 10\% FLOPs and maintained the same accuracy.

Inspired by CSPResNet, we proposed E-ConvNeXt by combining CSPNet with ConvNeXt. To further reduce the parameters and FLOPs, we adjust the channel numbers for all convolutions in the block. The number of channels in the 7×7 depthwise convolution and the MLP layer is halved. As shown in Figure \ref{fig:CSPduibi}, the number of channels in E-ConvNeXt is reduced from 96 to 48 and from 384 to 192 compared to ConvNeXt.
The formula for calculating depthwise convolution is
\begin{equation}
FLOPs=H_{out} \times W_{out} \times K^2 \times C_{out},
  \label{eq:FLOPs_DWConv}
\end{equation}
The FLOPs of ConNeXt Block are
\begin{align*}
    ((56 \times 56 \times 7^2 \times 96)\approx 15M)+ \\
    ((96 \times 56 \times 56 \times 1^2 \times 384) \approx 116M)+ \\
     ((384 \times 56 \times 56 \times 1^2 \times 96) \approx 116M)\approx 257M.
\end{align*}
The FLOPs of CSPConNeXt Block are
\begin{align*}
    ((56 \times 56 \times 7^2 \times 58)\approx 7.5M)+ \\
    ((48 \times 56 \times 56 \times 1^2 \times 192) \approx 29.5M)+ \\
     ((384 \times 56 \times 56 \times 1^2 \times 96) \approx 29.5M)\approx 66.5M.
\end{align*}
Obviously, by introducing the CSP Module, the FLOPs of the 
 original ConvNeXt is reduced by 60\%.

\subsubsection{Introducing transition hyperparameter}\label{chap:mid_channel}

\begin{figure} 
	\centering
		\includegraphics[width=1.1\linewidth]{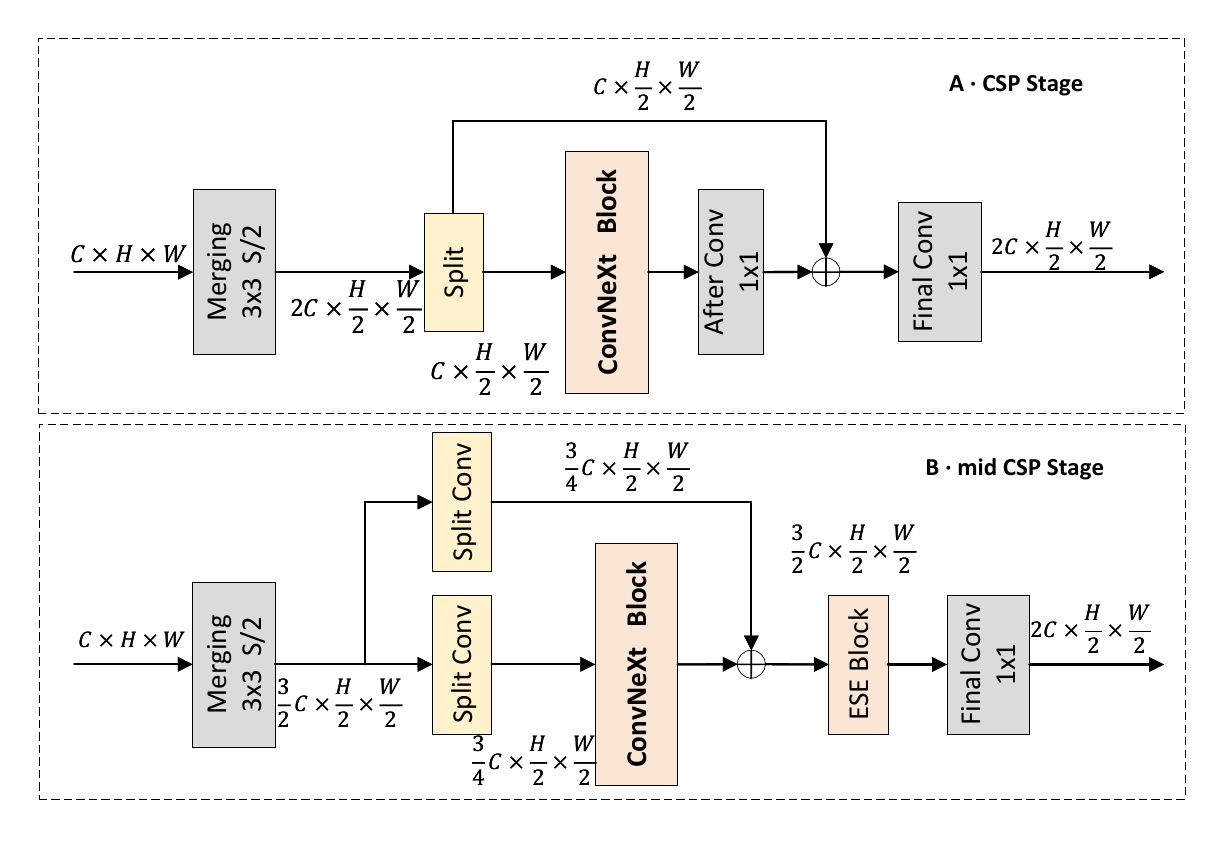}
	\caption{(A) is Original CSPConvNeXt 
CSP Stage. (B) is $ch_\text{mid}$ CSPConvNeXt 
 CSP Stage}
	\label{fig:Stage}
\end{figure}

 The structure of CSP Stage is shown in Figure \ref{fig:Stage} A. The CSP Stage uses splits the channels of the feature map into two parts via split operation. We use 1×1 convolution to replace the Split operation, which can better split the feature map.

The feature map channel transformation process of the original CSPNet is 
\[
ch_{\text{in}} \xrightarrow{\substack{\text{down-} \\ \text{sampling}}} ch_{\text{out}} \xrightarrow{\text{Split}} 2 \times (\frac{ch_{\text{out}}}{2})\xrightarrow{\text{concat}} ch_{\text{out}}
\] 
To further reduce the number of model parameters, we introduce an transition hyperparameter  $ch_\text{mid}$ . The formula for $ch_\text{mid}$ is $
ch_{\text{mid}} = \frac{ch_{\text{in}} + ch_{\text{out}}}{2}
$.
The new feature map channel  transformation process  is 
\[
ch_{\text{in}} \xrightarrow{\substack{\text{down-} \\ \text{sampling}}} ch_{\text{mid}} \xrightarrow{\substack{\text{1x1} \\ \text{Conv}}}  2 \times \frac{ch_{\text{mid}}}{2} \xrightarrow{\text{concat}} ch_{\text{mid}} \xrightarrow{\substack{\text{1x1} \\ \text{Conv}}} ch_{\text{out}}
\] 
We call this version as $ch_\text{mid}$ CSPConvNeXt. As shown in Figure \ref{fig:Stage} B, by introducing the $ch_\text{mid}$ hyperparameter, it further reduces the FLOPs by 40\% compared to the original CSPConvNeXt.

\subsubsection{Stage ratio}

In both the original CSPConvNeXt and the $ch_\text{mid}$ CSPConvNeXt, we have achieved a significant reduction in parameters and network complexity. However, as network complexity decreases, the model accuracy exhibits a corresponding decline. To achieve a balance between FLOPs and accuracy, we increase the FLOPs of the $ch_mid$ CSPConvNeXt to the same level as the Original CSPConvNeXt by adjusting the stage ratio. 
The adjusted stage ratio is as follows: input channels are [65, 128, 256, 512], output channels are [128, 256, 512, 1024], blocks are [3, 3, 9, 3].

\begin{figure*} 
	\centering
		\includegraphics[width=1\linewidth]{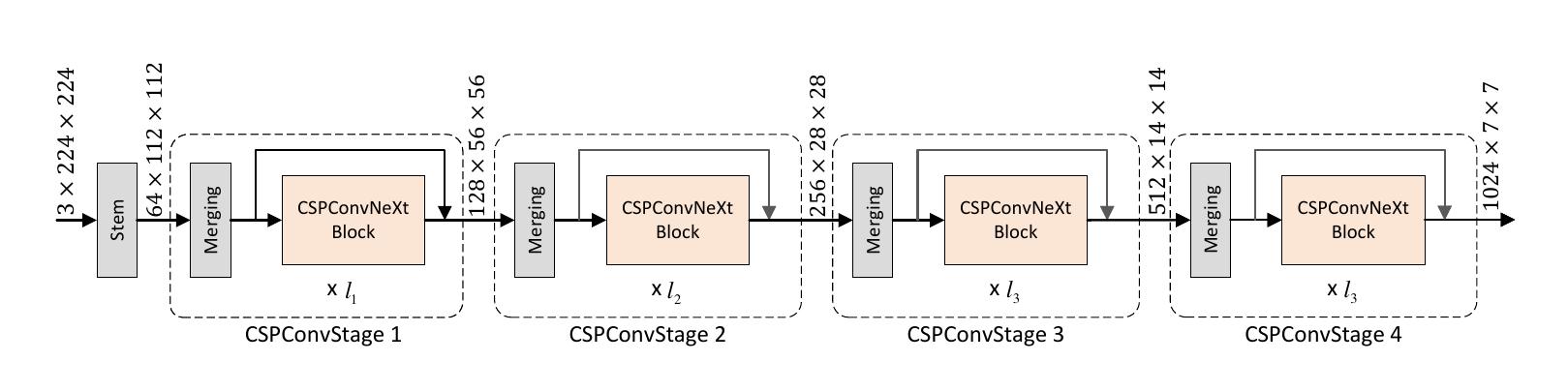}
	\caption{After adjusting the Stage ratio, CSPConvNeXt consists of one Stem and four CSPConvStages. Each Stage includes a downsampling layer and multiple CSPConvNeXt Blocks. The output of each Stage is twice the size of its input. The diagram shows the channel configuration of CSPConvNeXt-tiny.}
	\label{fig:hongguan}
\end{figure*}

The downsampling layers in ConvNeXt consist of four layers, namely one 4× downsampling layer followed by three 2× downsampling layers. To better adapt to the structure of CSP Module, the downsampling architecture of CSPConvNeXt is modified to five 2× downsampling layers, and each layer is composed of a 2x2 convolution with a stride of 2.
The final structure of CSPConvNeXt is shown in Figure \ref {fig:hongguan}. Compared with the original CSPConvNeXt, it has the same FLOPs but the accuracy increases 0.7\%.

\subsection{Optimization of Network Structure Design}

\subsubsection{Stepped Stem}

\begin{figure} 
	\centering
		\includegraphics[width=0.6\linewidth]{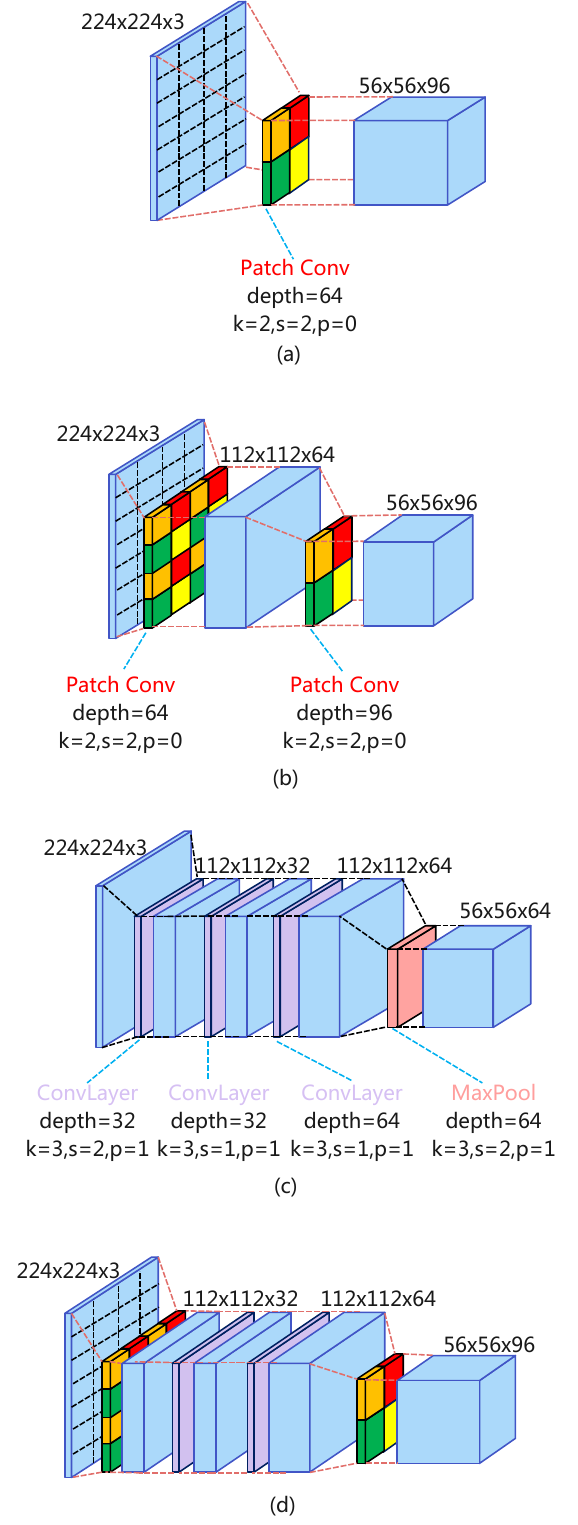}
	\caption{a is the Stem of ConvNeXt; b is the Stem of CSPConvNeXt after Stage ratio adjustment and the downsampling layer in the first CSPStage; c is the Stem layer of $ResNet_\text{vc}$; d is the Step Stem. The Step Stem is formed by integrating the characteristics of the Stem layers of ConvNeXt and $ResNet_\text{vc}$.}
	\label{fig:Stem}
\end{figure}

As shown in Figure \ref{fig:Stem} (a), the stem layer of ConvNeXt employs a 4×4 convolution with a stride of 4. The CSPConvNeXt stem structure is modified by replacing the single 4×4 convolution with two successive 2×2 convolutions with a stride of 2, the specific structure is shown in Figure \ref{fig:Stem} (b).
The first two downsampling operations in CSPConvNeXt will not only adjust the size of the feature map but also frequently modify the number of channels. 
This process will reduce the network accuracy.

Drawing inspiration from the designs of $ResNet_\text{vc}$ (as shown in Figure \ref{fig:Stem} c) and ConvNeXt, the stem layer of E-ConvNeXt is modified by replacing the original 2×2 convolution with a stepped convolution combination. This combination can resolve the above issue. It consists of one 2×2 convolution and two 3×3 convolutions, with the number of channels changing from 3 to 32 and then to 64. Specifically, the 2×2 convolution retains the Patchify concept. The two 3×3 convolutions enable better feature extraction. The stepped channel adjustment further mitigates the loss of spatial features.
The final stem structure is illustrated in Figure \ref{fig:Stem} d, which is called the Stepped Stem.

\subsubsection{Layer Normalization $\rightarrow$ Batch Normalization} 

ConvNeXt employs LN instead of BN. There are two modes of LN in ConvNeXt. Within the blocks, LN adopts the channel last mode, which is faster but requires transposing the feature maps. On the other hand, LN uses the channel first mode, which does not need feature map transposition but its speed is relatively slow. However, CSP Module requires a large number of convolution operations to adjust the number of channels, and the use of LN will seriously affect the running speed of the model.

\begin{table}[width=.9\linewidth,cols=3,pos=h]
    \caption{Comparison of efficiency between LN and BN: $Downsample1$ employs a 2×2 convolution with a stride of 2, followed by LN. $Downsample2$ utilizes a 2×2 convolution with a stride of 2, followed by BN. The input feature map has a size of [8, 64, 56, 56]. Repeating 10,000 forward and backward propagation processes}\label{tab:down_com}
    \begin{tabular*}{\tblwidth}{@{} LLL@{} }
        \toprule
        Name & DownSample-1 & DownSample-2\\
        \midrule
        Deploy-time & 16.41s & 29.65s \\
        Percentage & 55.3\% & 100\% \\
        \bottomrule
    \end{tabular*}
\end{table}

We tested LN and BN in the channel first mode. As shown in Table \ref{tab:down_com}, we designed two downsampling configurations for comparison. The results show that the downsampling operation using BN runs significantly faster than that using LN. Therefore, all LN in non-block ConvLayers are replaced with BN, and all non-block ConvLayers are converted into "Convolution + BN + GELU".

\begin{figure} 
	\centering
		\includegraphics[width=0.8\linewidth]{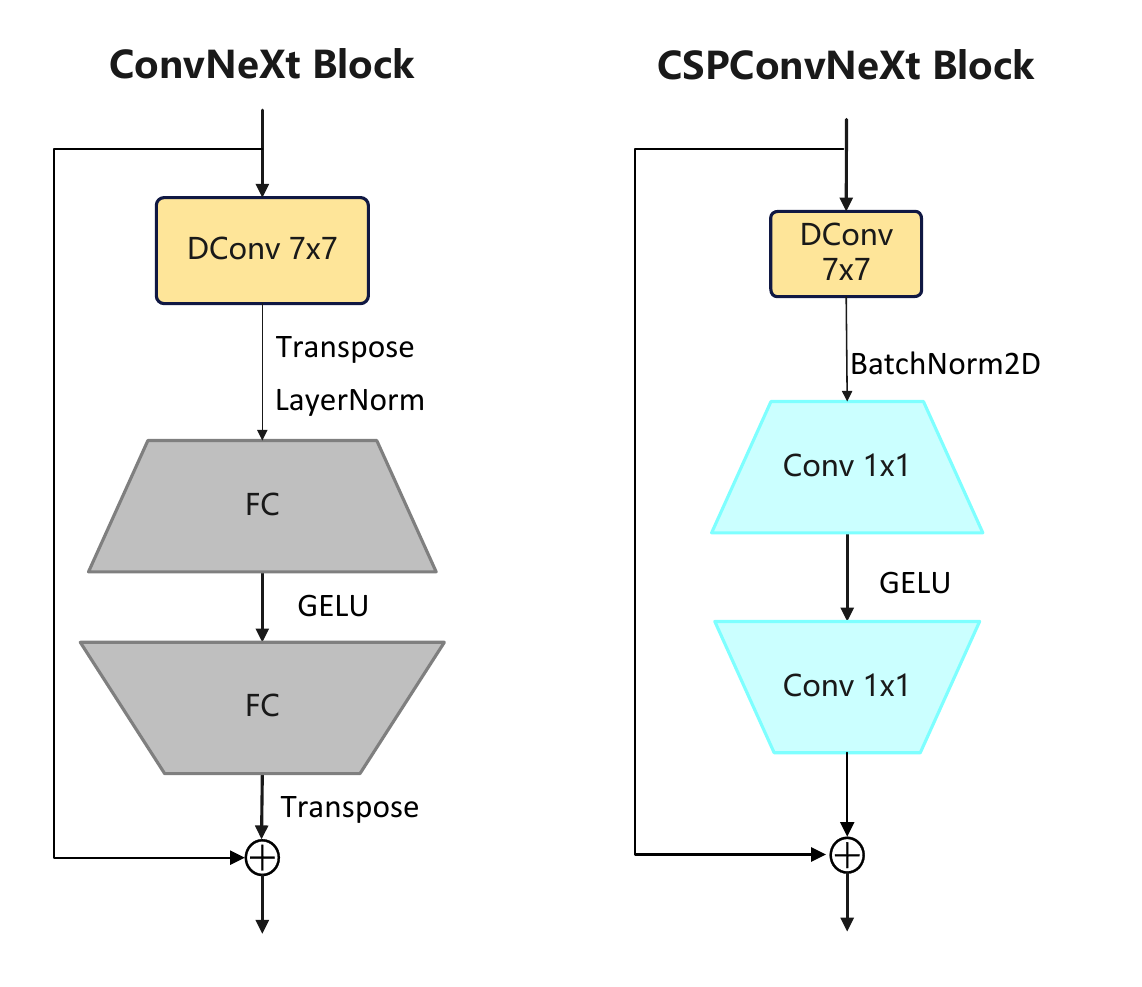}
	\caption{Comparison diagram of Block structures: The ConvNeXt Block uses fully connected layers to form the MLP layer, with Layer Normalization (LN) for normalization. The CSPConvNeXt Block employs 1x1 convolutions to construct the MLP layer, and adopts Batch Normalization (BN) for normalization.}
	\label{fig:compare_blocks}
\end{figure}

As shown in Figure \ref{fig:compare_blocks} (a), the LN in the ConvNeXt Block uses the channel last mode, which avoids inefficiency. However, this requires two transpose operations in the block layer to transpose the feature map, and also needs to use Fully Connected layers instead of 1x1 convolutions.
Thus, we replaced LN with BN in the block layer, removed transpose operations, and used 1x1 convolutions instead of Fully Connected Layers. The modified block structure is shown in Figure \ref{fig:compare_blocks} (b).
The modified block structure does not change the network's FLOPs or parameters, but increases the network speed by 20\% and brings a slight improvement in accuracy.

\subsection{Introducing Channel Attention}

The original ConvNeXt architecture uses LayerScale\cite{vaillant1994original} for feature scaling. In this paper, we replace this module with channel attention module, aiming to improve the accuracy of network. However, if channel attention is directly added to the end of the block, it will lead to training instability or model collapse.

\begin{figure} 
	\centering
		\includegraphics[width=1\linewidth]{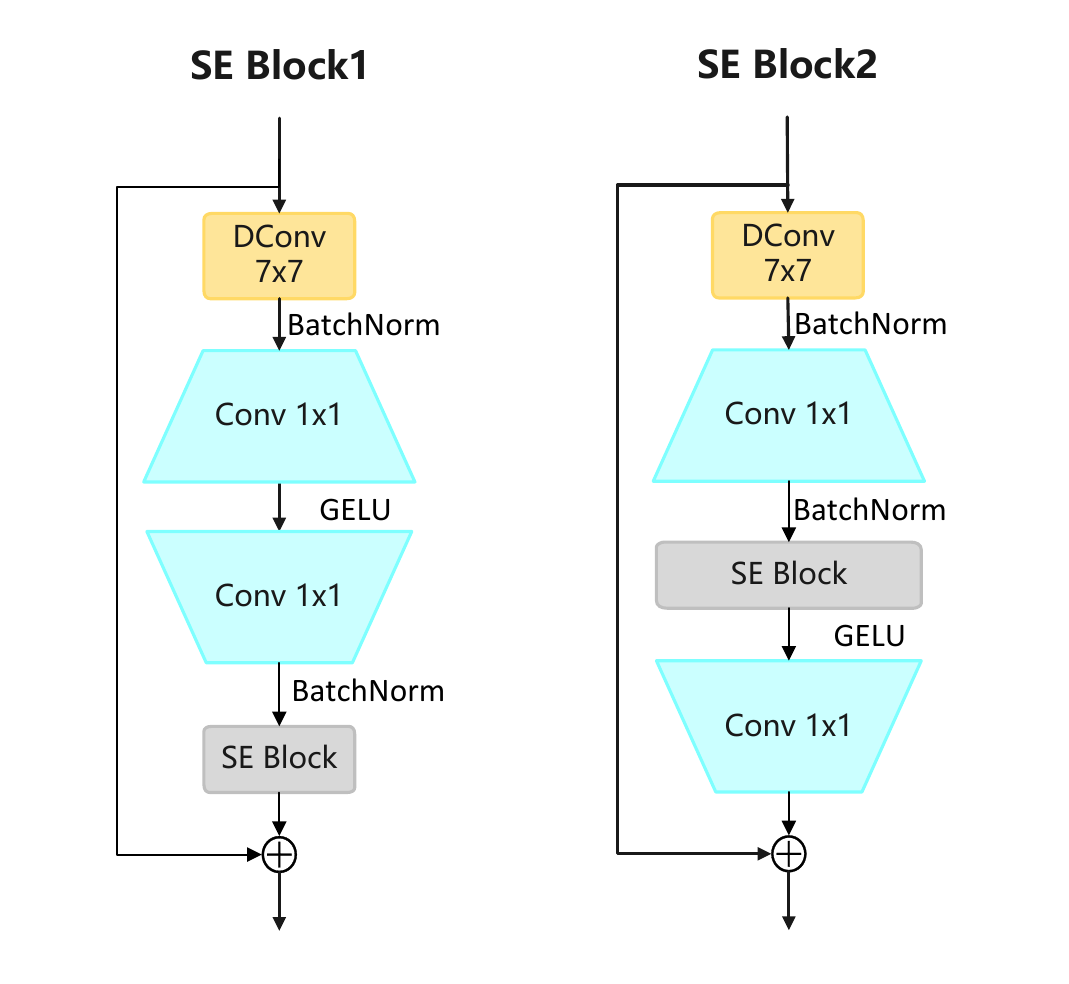}
	\caption{SE Block1 adopts the integration method of SENet, with the number of channels in the SE Block being c. SE Block2 follows the integration method of EfficientNet, where the number of channels in the SE Block is 4c.}
	\label{fig:zhuyili}
\end{figure}

To address the above issue, a normalization layer is introduced before the channel attention module. 
Referring the same method of introducing channel attention module in SENet and EfficientNet, the SE Block is added into ConvNeXt. The specific approach is shown in Figure \ref{fig:zhuyili}.
From the perspective of parameters and network complexity: the channel numbers in the SENet-style configuration is $c$, while that in the EfficientNet-style configuration reaches as high as $4c$ . Although both result in a similar increase in FLOPs, the channel expansion in EfficientNet introduces a substantial number of additional parameters, significantly boosting the model's parameters. Consequently, the integration style of SENet is more aligned with the design goal of E-ConvNeXt in maintaining lightweight characteristics. We incorporate the ESE Block into the network as a channel attention module.
After comparison, we integrate the ESE module into E-ConvNeXt.

\subsection{The overall architecture of E-ConvNeXt}\label{chap:CSPConvNeXt}

\begin{figure*} 
	\centering
		\includegraphics[width=1\linewidth]{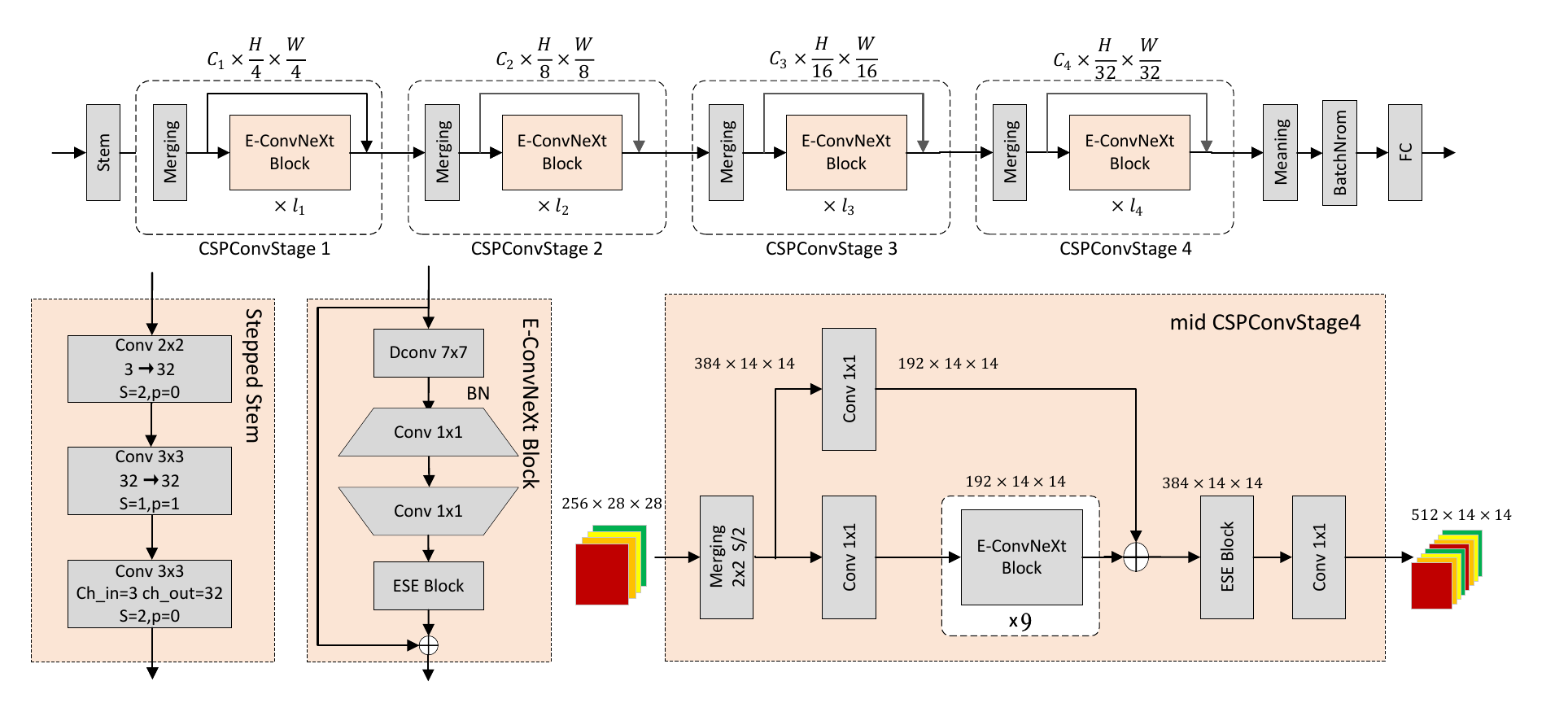}
	\caption{The overall architecture of E-ConvNeXt, illustrating the hierarchical stages, including the Stepped Stem, CSPConvStages, and the integration of key components such as ESE Block, Batch Normalization, and depthwise convolutions.}
	\label{fig:CSPConvNeXt}
\end{figure*}

By combining CSPNet with ConvNeXt and improving the network architecture, E-ConvNeXt is proposed for balancing accuracy, FLOPs and parameters. Figure \ref{fig:CSPConvNeXt} shows the overall architecture. It has five hierarchical stages,each stage is preceded by a $2 \times 2$ convolution with stride 2 for spatial downsampling and channel expansion.

 The first stage is the stem layer, which consists of a $2 \times 2$ convolution  with stride 2, followed by two regular $3 \times 3$ convolution with stride 1. The number of channels in the Stem layer follows a stepped pattern.
The following four stages are CSPConvStage. Each CSPConvStage contains a $2 \times 2$ convolution with stride 2 for downsampling, followed by splitting the feature map into two parts.ne feature map passes through the blocks and the other does not, and the feature maps are finally merging together. CSPConvStage introduces $ch_\text{mid}$ to further reduce the network's parameters and FLOPs. 1x1 conv is used to replace the Split operation.

Each CSPConvStage contains multiple Conv blocks. Each block consists of a  7x7 Dpethwise Conv followed by two  1x1 Conv layers. Together, they form an inverted residual block. We replace LN with BN, significantly improving the network 
 effective.  As for the activation layer, we keep use GELU. The final two layers are global average pooling and a fully connected layer, used for feature transformation and classification.

To meet a wide range of application requirements under different computational budgets, we present three variants of E-ConvNeXt. We call them CSPConvNeXt-mini, E-ConvNeXt-tiny, and E-ConvNeXt-Small respectively.They share a similar architecture but differ in depth and width.

\section{Experiments}

This paper used two benchmark datasets: ImageNet1K\cite{russakovsky2015imagenet} and ImageNet-100\footnote{https://huggingface.co/datasets/ilee0022/ImageNet100}.

\begin{itemize}
    \item ImageNet-1K: It is widely used in image classification tasks. It consists of 1,000 classes and contains approximately 1.28 million training images and 50,000 validation images. Each class represents a distinct category, ensuring the diversity and robustness of model evaluation.   
    \item ImageNet-100: It is a subset of ImageNet-1K, containing 100 classes. In is paper, it is primarily used to quickly verify the impact of different structures on model performance.
\end{itemize}

\subsection{Settings}

\begin{table}
\caption{\textbf{ImageNet-1k/100 (pre-)training settings.} The experimental setup of E-ConvNeXt on ImageNet-1k and ImageNet-100, for more detailed settings, please refer to the github repository.}
\centering
\scriptsize 
\label{tab:pretarining}
\begin{tabular}{>{\centering\arraybackslash}p{3cm} >{\centering\arraybackslash}p{2cm} >{\centering\arraybackslash}p{2cm}} 
\toprule
\textbf{(pre-)training config} & \textbf{\parbox[c]{5cm}{E-ConvNeXt \\ -M/T/S \\ ImageNet-1K \\ 224$^2$}} & \textbf{\parbox[c]{5cm}{E-ConvNeXt \\ -M/T/S \\ ImageNet-100 \\ 224$^2$}} \\ \midrule
Weight init                    & trunc. normal (0.2)                            & trunc. normal (0.2)                                \\
Optimizer                      & AdamW\cite{loshchilov2017decoupled}                                          & AdamW\cite{loshchilov2017decoupled}                                              \\
Base learning rate             & 1.25e-4 each 128 batch size                      & 1.25e-4 each 128 batch size                                            \\
Weight decay                   & 0.05                                           & 0.05                                               \\
Optimizer momentum             & $\beta_1, \beta_2=0.9, 0.999$                  & $\beta_1, \beta_2=0.9, 0.999$                      \\                                  
Training epochs                & 300                                            & 100                                                 \\
Learning rate schedule         & cosine decay                                   & cosine decay                                       \\
Warmup epochs                  & 20                                             & 5                                                 \\
Warmup schedule                & linear                                         & linear                                             \\
Layer-wise lr decay            & None                                           & None                                               \\
Randaugment\cite{cubuk2020randaugment}                   & (9, 0.5)                                       & (9, 0.5)                                           \\
Mixup\cite{zhang2017mixup}                          & 0.8                                            & 0.8                                                \\
Cutmix\cite{yun2019cutmix}                         & 1.0                                            & 1.0                                                \\
Random erasing\cite{zhong2020random}                 & 0.25                                           & 0.25                                               \\
Label smoothing\cite{szegedy2016rethinking}                & 0.1                                            & 0.1                                                \\
Head init scale                & None                                           & None                                               \\
Gradient clip                  & None                                           & None                                               \\                                 \bottomrule
\end{tabular}
\label{tab:shiyanshezhi}
\end{table}

We provide the pretraining settings of E-ConvNeXts on ImageNet-1K and ImageNet-100 in Table \ref{tab:shiyanshezhi}. All E-ConvNeXt variants use the same settings, except that the base learning rate increases proportionally with the batch size. For example, when training E-ConvNeXt-tiny on 4 NVIDIA 4080 GPUs, the batch size is set to 256 and the base learning rate is set to 2.5e-4. In contrast, when training on 8 NVIDIA 4090 GPUs, the batch size is set to 1024 and the base learning rate is set to 1e-3.

\subsection{Comparison of different CSP architectures}

To compare the performance of various CSP structures, we train and validate each version of CSPConvNeXt on ImageNet-1K.

\begin{itemize}

    \item original CSPConvNeXt: it combines CSPNet with ConvNeXt while preserving the channel ratio between convolution and MLP in the Block of ConvNeXt.

    \item $ch_\text{mid}$ CSPConvNeXt: it uses 1×1 convolution to replace the Split operation and introduces the transition hyperparameter $ch_\text{mid}$.

    \item final CSPConvNeXt: it is obtained by adjusting the downsampling layer and stage ratio of $ch_\text{mid}$ CSPConvNeXt.

\end{itemize}

As shown in the table \ref{tab:com_csp}, The original CSPConvNeXt can reduce FLOPs from 4.5G to 1.5G. The $ch_\text{mid}$ CSPConvNeXt can further reduce FLOPs to 0.9G. However, as FLOPs decrease, the accuracy of the models also shows a downward trend.

To further balance FLOPs and Accuracy, we adjusted the stage ratio of the $ch_\text{mid}$  CSPConvNeXt, resulting in the final CSPConvNeXt. The FLOPs of the final CSPConvNeXt are increased to the same level as the original CSPConvNeXt, but its accuracy is 0.7\% higher. Therefore, we selected the final CSPConvNeXt as the base model for subsequent experiments.

\begin{table}[width=.9\linewidth,cols=4]
   \caption{\textbf{Compare different CSP structures}}\label{tab:com_csp}
   \begin{tabular*}{\tblwidth}{@{} ccc@{} }
      \toprule
        & Top-1 Acc & FLOPs \\ \hline
     { original CSPConvNeXt } & 78.5 & 1.5G \\ 
     {$ch_\text{mid}$ CSPConvNeXt} & 77.8 & 0.9G \\
     {final CSPConvNeXt }  & 79.2 & 1.5G \\

     \bottomrule
   \end{tabular*}
   \end{table}

\subsection{Comparison of different Stem structures}

To verify whether the Stepped Stem can alleviate the feature loss caused by continuous downsampling, we compare the performance with the original Stem of CSPConvNeXt and the $ResNet_{vc}$ Stem. The dataset used in this experiment is ImageNet-1k, with the image size fixed at 224x224.

\begin{table}
\caption{\textbf{Compare different Stem structures}}
\setlength{\tabcolsep}{16pt}
\begin{tabular}{Lcc}
\toprule
 & TOP-1 ACC & FLOPs\\
 
\midrule
CSPConvNeXt Stem & 79.2 & 1.5G\\
$ResNet_\text{vc} $ Stem & 79.7 & 2.0G\\
Stepped Stem & 79.8 & 2.0G\\
\bottomrule
\end{tabular}
\label{tab:stem}
\end{table}

As shown in Table \ref{tab:stem}, compared with the original Stem of CSPConvNeXt, the $ResNet_{vc}$ Stem improved accuracy by 0.5\% while the FLOPs increased 0.5G . We further combined the Patchify Stem with the $ResNet_{vc}$ Stem to obtain the Stepped Stem. It further improved accuracy by 0.1\% with the FLOPs unchanged. Finally, we selected the Stepped Stem as the Stem of the network.

\subsection{Comparison of LN and BN}

To verify which normalization layer is more suitable for the downsampling layer of E-ConvNeXt, we replaced LN in the downsampling layer with BN. As shown in Table \ref{tab:downsample}, the network using BN achieves higher accuracy compared to that using LN. Additionally, BN operates more efficiently than LN in the channel first mode. Therefore, BN is selected as the normalization layer for the downsampling layer of CSPConvNeXt. 

\begin{table}
\caption{\textbf{Compare different Downsampling structure}}\label{tab:downsample}
\setlength{\tabcolsep}{8pt}
\begin{tabular}{lcc}
\toprule
 & \parbox{2.5cm}{\centering Downsample \\ LayerNorm} 
 & \parbox{2.5cm}{\centering Downsample \\ BatchNorm}\\
\midrule
Top-1 Acc & 79.8 & 79.9 \\
FLOPs & 2.0G & 2.0G \\
\bottomrule
\end{tabular}
\end{table}

To ensure network consistency, we replaced the LN with the BN in block. As shown in Table \ref{tab:Block_com}, compared to the Block using LN, the Block using BN achieved higher accuracy and 25\% speedup without altering the network's computational complexity. This indicates that BN is more suitable for E-ConvNeXt.

\begin{table}[width=1\linewidth]
\caption{\textbf{Compare with different block}}\label{tab:Block_com}
\begin{tabular*}{\tblwidth}{@{} Lcc@{} }
    \toprule
    {Evaluation Metrics}
     & { LN Block }
     & { BN Block }\\
    \midrule
    Top-1 Acc & 79.9 & 80.1 \\
    FLOPs & 2.0G & 2.0G \\
    FPS & 447 & 549 \\
    \bottomrule
\end{tabular*}
\end{table}

\subsection {Comparison of Different 
Channel Attention Modules}\label{chap:compare_attn}

We integrated various mainstream channel attention modules (namely SE, CBAM, ECA and ESE) into the baseline model and compared their performances. ImageNet-100 dataset is used in this experiment. The total number of training epochs is set to 60. The experimental results are shown in Table \ref {tab:which_attention}. Among all tested modules, ESE Block proformed the best. It achieved the highest accuracy and fastest inference time. Therefore, we add the ESE Block to E-ConvNeXt.

\begin{table} [width=1\linewidth,cols=6,pos=h]
\caption{Copare with different channel attention}
\setlength{\tabcolsep}{5pt}
\begin{tabular*}{\tblwidth}{@{} Lccccc@{} }
\toprule
 & None & SE Moudle & ECA & CBAM & ESE Layer\\
\midrule
Acc  & 82.1 & 83.4 & 84.2 & 83.9 & 85.4 \\
Time & 1.82ms & 19.3ms & 19.3ms & 2.05ms & 1.89ms \\
\bottomrule
\label{tab:which_attention}
\end{tabular*}
\end{table}

\subsection{Comparison of state-of-the-arts} \label{chap:compared_with_others}

To evaluate the proformence of E-ConvNeXt, we conducted a series of experiments on the ImageNet-1k. 
We compared our model with mainstream models such as ConvNeXt, StarNet, FasterNet, and Swin-Transformer.
To evaluate the performance of E-ConvNeXt under different network complexities, we compared E-ConvNeXt-mini, E-ConvNeXt-tiny, and E-ConvNeXt-small with other models of the same FLOPs, respectively.
The experimental results are shown in Table \ref{tab:imagenet}, where the parameters, FLOPs, and ImageNet-1K Top1-ACC of other methods are all derived from relevant references.

\begin{table}[width=1\linewidth,cols=4,pos=h]
  \caption{\textbf{ Comparison on ImageNet-1k benchmark.} Models with similar FLOPs are grouped together. For each group, our
E-ConvNeXt achieves the optimal balance between accuracy and FLOPs.}\label{tab:imagenet}
  \begin{tabular*}{\tblwidth}{@{} Lccc@{} }
    \toprule
    \cr model & \#param. & FLOPs & \begin{tabular}[c]{@{}c@{}} IN-1K \\ top-1 acc.\end{tabular} \\
    \hline
    {GhostNet $\times$1.3 \cite{han2020ghostnet}} & 
    {7.4M} &
    {0.24G} &
    {75.7}  \\
    {ShuffleNetV2 $\times$2 \cite{ma2018shufflenet}} &
      {7.4M} &
      {0.24G} &
    {75.7} \\
    {MobileNetV2 $\times$1.4 \cite{sandler2018mobilenetv2}} &
      {6.1M} &
      {0.60G} &
      {74.7} \\
    {MobileViT-XS \cite{mehta2021mobilevit}} &
      {2.3M} &
      {1.05G} &
      {74.8} \\
    {EdgeNeXt-XS \cite{maaz2022edgenext}} &
      {2.3M} &
      {0.54G} &
      {75.0} \\
    {FasterNet-T1 \cite{chen2023run}} &
      {7.6M} &
      {0.85G} &
      {76.2} \\
    {StarNet-S4 \cite{ma2024rewrite}} &
      {7.5M} &
      {1.07G} &
      {78.4} \\
    \rowcolor{gray!20}
    E-ConvNeXt-Mini (Ours)  & 7.6M & 0.93G  & 78.3 \\ 
    \midrule
    {CycleMLP-B1 \cite{chen2107cyclemlp}} &
      {15.2M} &
      {2.10G} &
      {79.1} \\
    {PoolFormer-S12 \cite{yu2022metaformer}} &
      {11.9M} &
      {1.82G} &
      {77.2} \\
    {MobileViT-S \cite{mehta2021mobilevit}} &
      {5.6M} &
      {2.03G} &
      {78.4} \\
    {EdgeNeXt-S \cite{maaz2022edgenext}} &
      {5.6M} &
      {1.26G} &
      {79.4} \\
    {FasterNet-T2 \cite{chen2023run}} &
      {15.0M} &
      {1.91G} &
      {78.9} \\ 
    \rowcolor{gray!20}
    E-ConvNeXt-Tiny (Ours)  & 13.2M & 2.04G  & 80.6 \\ \hline
    {ResNet50} &
      {25.6M} &
      {4.11G} &
      {78.8} \\
    {CycleMLP-B2 \cite{chen2107cyclemlp}} &
      {26.8M} &
      {3.90G} &
      {81.6} \\
    {PoolFormer-S24 \cite{yu2022metaformer}} &
      {21.4M} &
      {3.41G} &
      {80.3} \\
    {PoolFormer-S36 \cite{yu2022metaformer}} &
      {30.9M} &
      {5.00G} &
      {81.4} \\
    {PVT-Small \cite{wang2021pyramid}} &
      {24.5M} &
      {3.83G} &
      {79.8} \\
    {PVT-Medium \cite{wang2021pyramid}} &
      {44.2M} &
      {6.69G} &
      {81.2} \\
    {FasterNet-S \cite{maaz2022edgenext}} &
      {31.1M} &
      {4.56G} &
      {81.3} \\ 
    {Swin-T \cite{liu2021swin}} &
      {28.3M} &
      {4.51G} &
      {81.3} \\
    {ConvNeXt-T \cite{liu2022convnet}} &
      {28.6M} &
      {4.47G} &
      {82.1} \\
    \rowcolor{gray!20}
    E-ConvNeXt-Small (Ours)  & 19.4M & 3.12G  & 81.9 \\
    \bottomrule
  \end{tabular*}
  \end{table}

E-ConvNeXt achieves performance comparable to state-of-the-art models. Specifically, E-ConvNeXt-mini, with 0.93G FLOPs and 78.3\% accuracy, performed as well as the best model StarNet-S4.  Compared with networks such as FasterNet-T2 and MobileViT-S, E-ConvNeXt-tiny achieved the highest accuracy under similar FLOPs. It is able to balance FLOPs and accuracy. Compared with other networks of the same FLOPs, E-ConvNeXt-small belong to the first tier in terms of accuracy. Meanwhile, E-ConvNeXt-small has lower FLOPs.

\begin{figure}
    \centering
    \vspace{-0.05in}
    \includegraphics[width=1\linewidth]{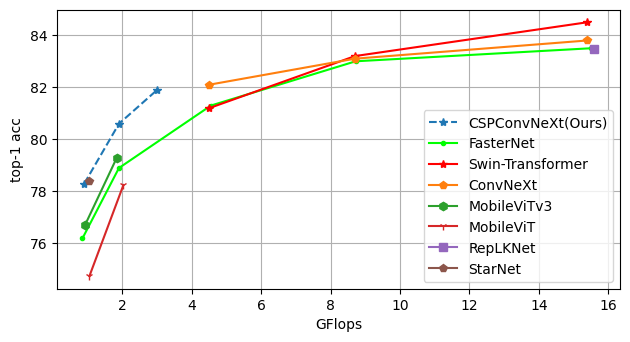}

    \vspace{-0.15in}
    \caption{\textbf{Rade-off curves figure.} E-ConvNeXt achieves the highest efficiency in balancing accuracy and FLOPs.}
    \label{fig:imagenet}
    \vspace{-0.15in}
\end{figure}

To further demonstrate the superiority of E-ConvNeXt, we plotted the data from Table \ref{tab:imagenet} into trade-off curves as shown in Figure \ref{fig:imagenet}. As indicated in Figure \ref{fig:imagenet}, E-ConvNeXt establishes a new state-of-the-art in balancing accuracy and FLOPs among all the networks evaluated.
The above results fully demonstrate the application prospects of E-ConvNeXt in lightweight scenarios.

\section{E-ConvNeXt on downstream tasks}

We conducted extensive experiments to verify the performance of E-ConvNeXt in object detection. The experiments were carried out on the sonar image dataset and the DUO dataset\cite{2021arXiv210605681L}.
In the experiments,we pretrained E-ConvNeXt on ImageNet-1K and  used it as the backbone network. It is then equipped with the PP-YOLOE\cite{xu2022pp} detector and YOLOv10\cite{wang2024yolov10} detector.

\subsection{Experiment Setup}

To be fairness, PP-YOLOE and YOLOv10 use the same hyperparameter settings: 36 training epochs, AdamW optimizer, base learning rate  3.125e-4, batch size 8, CosineDecay scheduler, and LinearWarmup strategy. All experiments are conducted on a server equipped with two NVIDIA GeForce 1080Ti GPUs (11GB memory) and the Ubuntu 20.04 operating system.

The data augmentation methods for PP-YOLOE and YOLOv10 follow those in their original papers: PP-YOLOE uses Random Crop, Random Horizontal Flip, Color Distortion, and Multi-scale; YOLOv10 uses Mosaic, Mixup, and Copy-paste.

\subsection{Evaluation Metrics}

In this experiment, mAP (mean Average Precision) is adopted as the core evaluation metric.
mAP is one of the most authoritative metrics for object detection. It is calculated based on Average Precision (AP), specifically as the arithmetic mean of the average precision values across all classes, with the calculation formula as follows:
\begin {equation}
mAP = \frac {1}{N} \sum_{i=1}^{N} AP_i
\end {equation} 
$N$ is the total number of classes in the dataset, and $AP_i$ denotes the average precision of the $i$-th class.

Average Precision (AP) is referring to the area under the Precision-Recall (P-R) curve for a single class. A higher AP value indicates better detection performance of the model on that class. Precision (P) and recall (R) are the core components of the P-R curve. The calculation formula of precision  as follows:
\begin {equation*}
P = \frac{TP}{TP + FP}
\end {equation*} 
where $TP$ (True Positive) denotes correctly detected positive samples, and $FP$ (False Positive) denotes negative samples incorrectly detected as positive ones.

The calculation formula as follows:
\begin {equation*}
R = \frac{TP}{TP + FN} 
\end {equation*} 
where $FN$ (False Negative) denotes actual positive samples that are not detected. Together, they form the vertical and horizontal axes of the P-R curve, serving as the core inputs for AP calculation.

\subsection{Object detection method for underwater sonar images} \label{chap:sonor_com}

In this section, we conduct extensive experiments to evaluate the effectiveness of E-ConvNeXt as the backbone network for PP-YOLOE and YOLOv10.

 We use the forward sonar data from Ocean Space Environment Awareness (Orca) open-source project\footnote{https://github.com/violetweir/Sonor\_dataset}. There are 5,000 images in total, of which 4,000 for training and 1,000 for test.Table \ref{tab:sonor_train} and \ref{tab:sonor_test} shows details about object categories and the number of sonar images for each category.

 \begin{table} [width=1\linewidth,cols=3,pos=h]
\caption{Sonor object information in training data}
\begin{tabular*}{\tblwidth}{@{} Lcc@{} }
\toprule
Object category & Number of objects &  Number of images \\

\midrule
Ball    &   1943 &  1941    \\
Circle cage     &   386 & 383 \\
Cube    &   1752 &  1749 \\
Cylinder        &   402 &   401 \\
Human body      &   684 &   683 \\
Metal   bucket  &   403 &   402 \\
Square cage     &   655 &   655 \\
Tyre            &   852 &   850 \\
ToTal           &   7077&   4000    \\
\bottomrule
\label{tab:sonor_train}
\end{tabular*}
\end{table}

\begin{table} [width=1\linewidth,cols=3,pos=h]
\caption{Sonor object information in training data}
\begin{tabular*}{\tblwidth}{@{} Lcc@{} }
\toprule
Object category & Number of objects & Number of images \\

\midrule
Ball    &   595 &  595    \\
Circle cage     &   86 & 86 \\
Cube    &   424 &  423 \\
Cylinder        &   39 &   39 \\
Human body      &   379 &   379 \\
Metal   bucket  &   44 &   43 \\
Square cage     &   169 &   169 \\
Tyre            &   108 &   108 \\
ToTal           &   1844&   1000   \\
\bottomrule
\label{tab:sonor_test}
\end{tabular*}
\end{table}

 \subsubsection{Results}

 \begin{table}[width=1\linewidth,cols=6,pos=h]
    \caption{Performance comparison of different backbones (including PaNet, ConvNeXt-T, and E-ConvNeXt-T) integrated with YOLOv10-L and YOLOv10-M on the sonar image dataset, evaluated by metrics: FLOPs, FPS, mAP, $mAP^m$ (medium-sized objects), and $mAP^l$ (large-sized objects).}\label{tab:sonor_yolov10}
\begin{tabular*}{\tblwidth}{@{} cccccc@{} }
\toprule
Backbone & 
FLOPs & 
FPS & 
$\text{mAP}$ & 
$\text{mAP}^{\text{m}}$ & 
$\text{mAP}^{\text{l}}$  \\
 \toprule

\begin{tabular}[c]{c}\small YOLOv10-L \\ {PaNet} \end{tabular}&
{120.3} &
{22.4} &
{46.1} &
{46.0} &
{41.5} 
\\
\rowcolor{gray!20}
\begin{tabular}[l]{c}\small YOLOv10-M \\ {ConvNeXt-T} \end{tabular}&
{94.3} &
{23.1} &
{50.1} &
{50.2} &
{32.8} 
\\
\begin{tabular}[c]{c}\small YOLOv10-M \\ {PaNet} \end{tabular}&
{59.1} &
{28.2} &
{45.6} &
{45.5} &
{43.6} 
\\
\rowcolor{gray!20}
\begin{tabular}[l]{c}\small YOLOv10-M \\ {E-ConvNeXt-T} \end{tabular}&
{69.9} &
{26.9} &
{51.3} &
{51.4} &
{39.0} 
\\

\toprule

\bottomrule
\end{tabular*}
\end{table}

To compare the performance of networks in different FLOPs, we selected YOLOv10-L and YOLOv10-M. As shown in Table \ref{tab:sonor_yolov10}, although YOLOv10-L has 0.5\% higher mAP than YOLOv10-M, YOLOv10-L’s FLOPs are 103\% higher than YOLOv10-M, and the large increase in FLOPs only brings a slight improvement in accuracy. When the backbone of YOLOv10-M is replaced with ConvNeXt-Tiny and E-ConvNeXt-Tiny, the accuracy of YOLOv10 improves compared to the original YOLOv10-L and YOLOv10-M.

Among the above models, E-ConvNeXt-tiny as the backbone of YOLOv10 achieves the highest mAP with the small FLOPs. The mAP of YOLOv10 is significantly improved after being equipped with E-ConvNeXt-tiny and ConvNeXt-tiny. This demonstrates the effectiveness of the ConvNeXt structure. Meanwhile, YOLOv10 with E-ConvNeXt-tiny achieves a higher mAP and lower FLOPs compared to YOLOv10 with ConvNeXt-tiny. This further illustrates the effectiveness of E-ConvNeXt as a backbone in the field of sonar images.

\begin{table}[width=1\linewidth,cols=6,pos=h]
    \caption{Performance comparison of different backbones (including E-ConvNeXt-Mini, E-ConvNeXt-Tint) integrated with PPYOLOE-S and PPYOLOE-L on the sonar image dataset, evaluated by metrics: FLOPs, FPS, mAP, $mAP^m$ (medium-sized objects), and $mAP^l$ (large-sized objects).}\label{tab:sonor_ppyoloe}
\begin{tabular*}{\tblwidth}{@{} cccccc@{} }
\toprule
Backbone & 
FLOPs & 
FPS & 
$\text{mAP}$ & 
$\text{mAP}^{\text{m}}$ & 
$\text{mAP}^{\text{l}}$  \\
 \toprule

\begin{tabular}[c]{c}\small PPYOLOE-S  \end{tabular}&
{17.4} &
{46.0} &
{42.6} &
{42.0} &
{28.1} \\

\rowcolor{gray!20}
\begin{tabular}[c]{c}\small PPYOLOE-S \\ E-ConvNeXt- \\Mini  \end{tabular}&
{20.56} &
{43.2} &
{50.6} &
{50.8} &
{44.8} \\

\begin{tabular}[c]{c}\small PPYOLOE-L  \end{tabular}&
{110.7} &
{33.8} &
{49.0} &
{48.8} &
{35.5} \\

\rowcolor{gray!20}
\begin{tabular}[c]{c}\small PPYOLOE-L \\ E-ConvNeXt- \\Tiny  \end{tabular}&
{82.6} &
{36.3} &
{51.3} &
{51.3} &
{44.4} \\
\toprule

\bottomrule
\end{tabular*}
\end{table}

To compare the performance of networks in different FLOPs, we selected two versions of PP-YOLOE: PP-YOLOE-S and PPYOLO-L. As shown in Table \ref{tab:sonor_ppyoloe}, the mAP steadily increases as the network complexity from small to large. By comparing  the performances in terms of $mAP^l$ and $mAP^m$, we find that PP-YOLOE behaves poorly in recognizing large target objects

 When using E-ConvNeXt-Mini as the backbone for PP-YOLOE-S, the mAP reached 50.6 and $mAP^L$ was 44.8. When using CSPConvNeXt-Tiny as the backbone for PP-YOLOE-L, the mAP reached 51.3 and $mAP^L$ was 44.4. Compared with the original PP-YOLOE, PP-YOLOE with E-ConvNeXt backbone shows significant improvements in sonar image object detection, especially fot large target objects.

 \subsection{Object detection for underwater optical images}

 In Section \ref{chap:sonor_com}, compared with PP-YOLOE, YOLOv10 with E-ConvNeXt achieves better performance. Therefore, we use YOLOv10 with E-ConvNeXt in this experiment.

 Detecting Underwater Objects (DUO)\footnote{https://github.com/chongweiliu/DUO} dataset contains 15632 optical images. It is a typical object detection dataset which has been widely used. As shown in Table \ref{tab:duo_train} and Table \ref{tab:duo_test}, DUO includes 74903 annotated objects across four common categories. Therefore, we select DUO as this section dataset.

\begin{table} [width=1\linewidth,cols=3,pos=h]
\caption{Object information in training data}
\begin{tabular*}{\tblwidth}{@{} Lcc@{} }
\toprule
Object category & Number of objects &  Number of images \\

\midrule
holothurian    &   6808 &  3001    \\
echinus     &   42955 & 5872 \\
scallop    &   1707 &  487 \\
starfish        &   12528 &   4078 \\
total        &     63998 &  13438 \\
\bottomrule
\label{tab:duo_train}
\end{tabular*}
\end{table}

\begin{table} [width=1\linewidth,cols=3,pos=h]
\caption{Object information in test data}
\begin{tabular*}{\tblwidth}{@{} Lcc@{} }
\toprule
Object category & Number of objects & Number of images \\

\midrule
holothurian    &   1079 &  490    \\
echinus     &   7201 & 967 \\
scallop    &   217 &  78 \\
starfish        &   2020 &   659 \\
total    &   10517   &   2194 \\
\bottomrule
\label{tab:duo_test}
\end{tabular*}
\end{table}

\subsubsection{Results }

\begin{table}
\caption{Performance comparison of different backbones (PaNet and E-ConvNeXt-T) with YOLOv10-L and YOLOv10-M on the COCO underwater dataset, including metrics of FLOPs, FPS, mAP, $mAP^m$ (medium object mAP), and $mAP^l$ (large object mAP).}
\begin{tabular*}{\tblwidth}{@{} cccccc@{} }
\toprule
Backbone & 
FLOPs & 
FPS & 
$\text{mAP}$ & 
$\text{mAP}^{\text{m}}$ & 
$\text{mAP}^{\text{l}}$  \\
 \hline
\begin{tabular}[c]{c}\small YOLOv10-L \\ {PaNet} \end{tabular}&
{120.3} &
{22.4} &
{56.4} &
{40.7} &
{62.0} 
\\
\rowcolor{gray!20}
\begin{tabular}[l]{c}\small YOLOv10-L \\ {E-ConvNeXt-T} \end{tabular}&
{78.0} &
{25.4} &
{61.2} &
{49.7} &
{70.1} 
\\
\begin{tabular}[c]{c}\small YOLOv10-M \\ {PaNet} \end{tabular}&
{59.1} &
{28.2} &
{56.2} &
{41.9} &
{60.7} 
\\
\rowcolor{gray!20}
\begin{tabular}[l]{c}\small YOLov10-M \\ {E-ConvNeXt-T} \end{tabular}&
{36.7} &
{32.8} &
{60.1} &
{56.0} &
{67.0} 
\\

\toprule

\bottomrule
\end{tabular*}
\label{tab:COCO_underwater}
\end{table}

As shown in Table \ref{tab:COCO_underwater}, we selected two versions of YOLOv10: YOLOv10-L and YOLOv10-M. The FLOPs of large version is twice that of medium version, but the mAP does not differ significantly. This is mainly because the original YOLOv10 typically requires hundreds of training epochs. To be fairness, all models were trained with 36 epochs in this experiment.

Using E-ConvNeXt as the backbone, YOLOv10 achieves better performance on underwater optical image detection task.  Specifically, the mAP of the YOLOv10-L with E-ConvNeXt-tiny is 61.2, which is 5.3 higher than that of the original YOLOv10-L. The mAP of the YOLOv10-M with E-ConvNeXt-tiny also shows an improvement compared with the original YOLOv10-M.  This result indicates that E-ConvNeXt has significant advantages in feature extraction tasks for optical images, and it is particularly suitable for lightweight application scenarios such as object detection tasks in complex optical environments.

\section{Conclusion}

 This paper presents the E-ConvNeXt framework, aiming to balance performance and efficiency, and address the limitations of existing ConvNeXt-based models in lightweight application scenarios.  The proposed integration of CSPNet with ConvNeXt, along with optimized Stem and Block structures and the introduction of the ESE Block, significantly reduces network complexity while enhancing feature expression capability and operational efficiency .  The training and testing strategies further improve performance and efficiency.  Extensive experiments on the ImageNet-1K dataset, as well as transfer learning tests on object detection tasks, validate the effectiveness, efficiency, and transferability of this method in lightweight scenarios and downstream tasks. Our work encourages further exploration across tasks, with codes available at \href{https://github.com/violetweir/E-ConvNeXt}{https://github.com/violetweir/E-ConvNeXt}

\textbf{Limitations and Future Work.} E-ConvNeXt, despite its advantages in balancing accuracy and efficiency, still has certain limitations. For instance, its performance in more extreme lightweight scenarios (e.g., ultra-low power consumption embedded devices) needs further verification and improvement. In the future, we will focus on optimizing the model structure to further reduce parameters and computational complexity while maintaining accuracy, to better adapt to extreme lightweight application scenarios.

\printcredits

\bibliographystyle{unsrt}

\bibliography{ref.bib}





\end{document}